\documentclass[sn-mathphys]{sn-jnl}

\usepackage{booktabs, tabularx}

\newcolumntype{C}{>{\centering\arraybackslash}X}

\usepackage{xcolor}

\jyear{2022}%

\theoremstyle{thmstyleone}%
\theoremstyle{thmstyletwo}%

\theoremstyle{thmstylethree}%

\begin{document}

\title[HMD-EgoPose: Marker-Less Pose Estimation for Surgical Guidance]{HMD-EgoPose: Head-Mounted Display-Based Egocentric Marker-Less Tool and Hand Pose Estimation for Augmented Surgical Guidance}


\author*[1,2]{\fnm{Mitchell} \sur{Doughty}}\email{mitchell.doughty@mail.utoronto.ca}

\author[1,2,3]{\fnm{Nilesh R.} \sur{Ghugre}}

\affil[1]{\orgdiv{Department of Medical Biophysics}, \orgname{University of Toronto}, \orgaddress{\city{Toronto}, \country{Canada}}}
\affil[2]{\orgdiv{Schulich Heart Program}, \orgname{Sunnybrook Health Sciences Centre}, \orgaddress{\city{Toronto}, \country{Canada}}}
\affil[3]{\orgdiv{Physical Sciences Platform}, \orgname{Sunnybrook Research Institute}, \orgaddress{\city{Toronto}, \country{Canada}}}




\abstract{\textbf{Purpose:} The success or failure of modern computer-assisted surgery procedures hinges on the precise six-degree-of-freedom (6DoF) position and orientation (pose) estimation of tracked instruments and tissue. In this paper, we present HMD-EgoPose, a single-shot learning-based approach to hand and object pose estimation and demonstrate state-of-the-art performance on a benchmark dataset for monocular red-green-blue (RGB) 6DoF marker-less hand and surgical instrument pose tracking. Further, we reveal the capacity of our HMD-EgoPose framework for performant 6DoF pose estimation on a commercially available optical see-through head-mounted display (OST-HMD) through a low-latency streaming approach. 
 
\textbf{Methods:} Our framework utilized an efficient convolutional neural network (CNN) backbone for multi-scale feature extraction and a set of subnetworks to jointly learn the 6DoF pose representation of the rigid surgical drill instrument and the grasping orientation of the hand of a user. To make our approach accessible to a commercially available OST-HMD, the Microsoft HoloLens 2, we created a pipeline for low-latency video and data communication with a high-performance computing workstation capable of optimized network inference.

\textbf{Results:} HMD-EgoPose outperformed current state-of-the-art approaches on a benchmark dataset for surgical tool pose estimation, achieving an average tool 3D vertex error of 11.0 mm on real data and furthering the progress towards a clinically viable marker-free tracking strategy. Through our low-latency streaming approach, we achieved a round trip latency of 199.1 ms for pose estimation and augmented visualization of the tracked model when integrated with the OST-HMD.

\textbf{Conclusion:} Our single-shot learned approach, which optimized 6DoF pose based on the joint interaction between the hand of a user and a rigid surgical drill, was robust to occlusion and complex surfaces and improved on current state-of-the-art approaches to marker-less tool and hand pose estimation. Further, we presented the feasibility of our approach for 6DoF object tracking on a commercially available OST-HMD.}




\maketitle

\section{Introduction}\label{sec1}
Augmented reality (AR) has been described as a potentially disruptive technology in the medical field due to its ability to enhance task localization and accuracy via the direct visualization of three-dimensional (3D) augmented virtual entities \cite{navab2012first}. 

In the context of state-of-the-art AR-led surgical navigation systems, continuous and precise intraoperative localization of surgical tools with respect to the patient anatomy is essential to the success of a procedure and safety of the patient \cite{sorriento2019optical}. Recent work has highlighted the potential benefits of optical see-through head-mounted displays (OST-HMDs) as the visualization medium for leading surgical navigation, as opposed to the typical monitor-led display of commercial surgical navigation suites \cite{doughty2022head}. Optical see-through HMDs allow for elimination of the visual disconnect present between the information presented on a monitor and the surgical scene \cite{muller2020augmented} and, using context-aware predictions from egocentric video, ensure that the current virtual augmentation meets the current information needs of the user \cite{doughty2021surgeonassist}.

Recent work has indicated the applicability of OST-HMDs for leading surgical navigation in laparoscopic and endoscopic procedures \cite{bernhardt2017status}, neurosurgery \cite{meola2017augmented}, orthopedic surgery \cite{jud2020applicability}, targeted cardiac procedures \cite{doughty2022head}, and general surgery \cite{rahman2020head}; however, there remains the concern of inconsistent intraprocedural instrument position and orientation (pose) estimation contributing to failed registration and tracking, and the potential for poor surgical outcomes \cite{fitzpatrick2010role}.  

\subsection{Related work}
Marker-free object pose estimation remains a challenging and important problem in the context of computer vision and AR. There is a substantial body of research work with proposed solutions to the problem of 6DoF object pose recovery; prior strategies have included template-based \cite{hinterstoisser2012model}, point-to-point \cite{drost2010model}, conventional learning-based \cite{brachmann2014learning}, and deep learning-based techniques \cite{sahin2018recovering}. Deep learning-led techniques are the current best-performing approaches in the 6DoF object pose estimation space and leverage large amounts of high-quality curated data to learn deep discriminative feature representations \cite{sahin2018recovering}. 

\subsubsection{Object pose estimation}
We can broadly categorize the deep learning-based 6DoF object pose estimation from monocular red-green-blue (RGB) literature into single-shot and iterative refinement approaches. Single-shot strategies have used a convolutional neural network (CNN) to predict 6DoF object pose directly without requiring multiple stages or hypothesis \cite{tekin2018real, Xiang-RSS-18, bukschat2020efficientpose}. Iterative refinement techniques have used a CNN to first predict the two-dimensional (2D) locations of the 3D bounding boxes which define an object in image space, then obtain the 6DoF object pose via perspective-n-point (PnP) or additional iterative refinement like random sample consensus (RANSAC) \cite{peng2019pvnet, song2020hybridpose, rad2017bb8}. Due to their indirect pose estimation approach, iterative refinement-based techniques typically report slower inference times and require more computational power than single-shot approaches \cite{bukschat2020efficientpose}.

\subsubsection{Hand pose estimation}
Prior approaches to hand pose estimation from monocular RGB data have focused on reducing the complexity of the problem by using a set of predefined hand poses or simplified hand representations \cite{athitsos2003estimating}. Several recent strategies have focused on directly regressing the 3D skeleton joint positions from RGB input frames \cite{cai2018weakly, mueller2018ganerated}, where others have used parametric hand models with shape and pose parameters to describe 3D hand mesh representations \cite{romero_embodied_2017}.

\subsubsection{Joint hand and object pose estimation}
Due to the nature of hand-object interactions, there are large mutual occlusions which occur during object handling and manipulation, making accurate pose estimation a difficult problem \cite{hasson2019learning}. In recent work, Hasson et al. built on the parametric hand model, MANO \cite{romero_embodied_2017}, and used a CNN-based approach (HandObjectNet) to leverage constraints imposed by typical hand-object interactions and reconstruct hand and object pose from monocular RGB video \cite{hasson2019learning, hasson2020leveraging}. Related to the surgical domain, Hein et al. recorded a monocular RGB-based benchmark dataset of hand and surgical tool interactions and investigated the performance and feasibility of different learned approaches to joint hand and object pose estimation \cite{hein2021towards}. 

\subsubsection{Objectives}
Our goals were to design a single-shot network which improved on the accuracy and speed of prior 6DoF rigid object and hand pose estimation techniques using monocular RGB data, and to demonstrate the feasibility of an OST-HMD led tracking experience using a streaming approach with egocentric video captured from the headset. 
With HMD-EgoPose, we demonstrated state-of-the-art performance on a benchmark dataset for surgical drill and hand pose estimation. Further, we revealed the feasibility of a low-latency streaming approach to send egocentric monocular RGB data to a high-performance computing workstation for optimized inference and enable 6DoF pose estimation on a commercially available OST-HMD. Our code is publicly available at \url{https://github.com/doughtmw/hmd-ego-pose}.

\section{Materials and Methods}\label{sec2}
\begin{figure}[h]%
    \centering
    \includegraphics[width=0.95\textwidth]{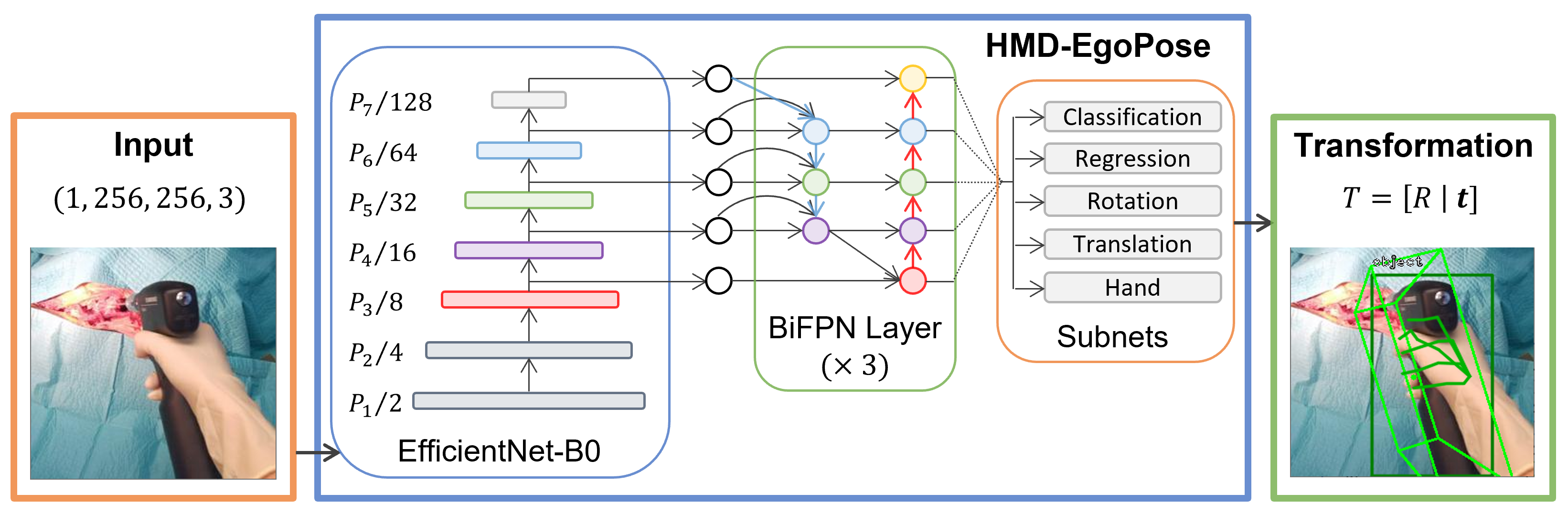}
    \caption{Overview of HMD-EgoPose, the single-shot deep learning-based framework for the prediction of relevant information regarding the joint hand and surgical tool representation from a single monocular red-green-blue (RGB) input image frame.}\label{fig1}
    \end{figure}

To jointly model the interactions between a surgical tool and the hand of a user, we built our network based on the EfficientDet-D0 \cite{tan2020efficientdet} backbone for multi-scale feature extraction and took influence from the EfficientPose \cite{bukschat2020efficientpose} architecture to introduce several subnetworks, which achieved the goal of lightweight single-shot 6DoF hand and tool pose estimation (Figure \ref{fig1}). 

Our selection of EfficientDet-D0 as the base framework for our pose detection strategy was informed by the results presented in EfficientPose and our goal of optimal network latency. When compared with EfficientDet-D0, Bukschat et al. reported a $3.1\%$ accuracy improvement with a larger EfficientDet-D3 backbone, at the cost of $2.91 \times$ slower model inference \cite{bukschat2020efficientpose}.

\subsection{Joint hand and tool estimation}
Using a CNN-based approach, we proposed HMD-EgoPose, an architecture for 6DoF rigid object and hand pose estimation based on learned deep discriminative features about joint hand-and-tool interactions. Our method takes as input monocular RGB video from an egocentric point of view that captures the hands of a surgeon interacting with a rigid surgical tool and provides an estimate of 6DoF pose of the surgical tool in frame and the 3D skeleton positions of hands. As with other marker-free strategies, the 3D geometry of the tracked object must be known during training, and an identical 3D model must be used for optimal  inference.

\subsubsection{Multiscale feature fusion}
To address scaling issues common to CNNs, Tan and Le have proposed the EfficientNet framework, generated from a compound scaling method which uniformly scales the network width, depth, and resolution based on a fixed set of scaling coefficients \cite{tan2019efficientnet}. Building off of the EfficientNet backbone, Tan et al. introduced EfficientDet, a network which incorporated a weighted bi-directional feature pyramid network (BiFPN) for multi-scale feature fusion to effectively represent and process multi-scale features for object detection \cite{tan2020efficientdet}. To extend the state-of-the-art 2D object detection performance of EfficientDet into our 3D task, we took influence from the EfficientPose \cite{bukschat2020efficientpose} architecture and introduced a series of subnetworks to estimate the 3D vertex positions of the hands of a user and 6DoF transformation (rotation and translation) describing the pose of the object in frame. As in similar object detection frameworks like YOLO \cite{redmon2016you}, EfficientDet does not directly predict 2D bounding boxes, but instead predicts the probabilities that correspond to tiled anchor boxes across an image, with a unique set of predictions given for each anchor box \cite{tan2020efficientdet}. In our network, the features contained within each anchor box serve as input to the subnetworks described.

\subsubsection{Rotation network}
Our rotation subnetwork used an axis-angle representation and directly regressed a rotation vector ($\boldsymbol{r} \in SO(3)$) from convolutional features extracted within each anchor box. To train our rotation subnetwork for axis-angle rotation prediction, we used the average squared distance between points of the correct model pose ($\boldsymbol{r}$) and their corresponding predictions ($\boldsymbol{\tilde{r}}$) \cite{Xiang-RSS-18}. We defined our rotation loss as:
\begin{equation}
\text{RLoss}(\boldsymbol{\tilde{r}}, \boldsymbol{r}) = \frac{1}{2m}\sum\limits_{x\in M}\|\tilde{\boldsymbol{r}}x-\boldsymbol{r}x\|^2\label{eq1}
\end{equation}
where $M$ is the set of 3D model points, $m$ is the number of points in the set, $\boldsymbol{r}$ denotes the axis-angle rotation for the ground truth pose, and $\tilde{\boldsymbol{r}}$ indicates the axis-angle rotation for the predicted pose.

\subsubsection{Translation network}
Taking influence from PoseCNN \cite{Xiang-RSS-18}, our translation subnetwork regressed a 2D object center point $\boldsymbol{c}=(c_x,c_y)^\intercal$ in pixel coordinates and a translation distance component $t_z$, separately, from convolutional features extracted within each anchor box. A translation vector ($\boldsymbol{t}\in \mathbb{R}^3$) for the object was then composed using the object center point $\boldsymbol{c}$, the distance component $t_z$, and knowledge of the camera intrinsic parameters. The missing translation components, $t_x$ and $t_y$ were computed as follows:
\begin{equation}
    t_x = \frac{(c_x - p_x) \cdot t_z}{f_x},\label{eq2}
    \end{equation}
\begin{equation}
    t_y = \frac{(c_y - p_y) \cdot t_z}{f_y}\label{eq3}
    \end{equation}
where the principal point $\boldsymbol{p}=(p_x,p_y )^\intercal$ and focal lengths $f_x$, $f_y$ were derived from the camera intrinsic parameters. We modeled our translation loss using the smoothed $\mathrm{L1}$ loss between the ground truth translation ($\boldsymbol{t}$) and the predicted translation ($\boldsymbol{\tilde{t}}$) as:
\begin{equation}
    \text{TLoss}(\tilde{\boldsymbol{t}}, \boldsymbol{t}) = \frac{1}{2m}\sum \limits_{x\in M} \mathrm{smooth_{L1}} (\boldsymbol{\tilde{t}}x - \boldsymbol{t}x),\label{eq4}
    \end{equation}
where
\begin{equation}
    \mathrm{smooth_{L1}} = \begin{cases}
        0.5x^2, & \text{if } \lvert x \rvert < 1, \\
        \lvert x \rvert - 0.5, & \text{otherwise }
    \end{cases}\label{eq5}
    \end{equation}
and where $M$ is the set of 3D model points, $m$ is the number of points in the set, $\boldsymbol{t}$ denotes the translation for the ground truth pose, and $\tilde{\boldsymbol{t}}$ indicates the translation for the predicted pose. The final 6DoF transformation, $\mathrm{T}$, for the predicted object pose was composed of the rotation matrix $\mathrm{R}\in \mathbb{R}^{3 \times 3}$, computed from the axis-angle rotation representation $\boldsymbol{r}$, and translation components as $\mathrm{T}=[\mathrm{R} \mid \boldsymbol{t}]  \in \mathbb{R}^{4 \times 4}$.

\subsubsection{Hand network}
Our hand subnetwork used a vector representation to model a 3D skeleton described by a total of 21 3D joint poses ($\boldsymbol{h} \in \mathbb{R}^{3 \times 21}$). The ground truth hand data used in training is in an identical vector format. Unlike other parametric approaches like MANO \cite{romero_embodied_2017} which generated a representative 3D hand mesh model from the 3D skeleton data, we instead directly regressed the 3D hand skeleton vector from convolutional features extracted from within each anchor box.

To train our hand subnetwork for the prediction of the 3D skeleton vertices of the hand, we used the average squared distance between points of the correct hand skeleton pose ($\boldsymbol{h}$) and their corresponding predictions ($\tilde{\boldsymbol{h}}$). We defined our hand vertex loss as:
\begin{equation}
    \text{HLoss}(\tilde{\boldsymbol{h}}, \boldsymbol{h}) = \frac{1}{2m}\sum \limits_{x\in M} \mathrm{smooth_{L1}} (\boldsymbol{\tilde{h}}x - \boldsymbol{h}x),\label{eq6}
    \end{equation}
where $M$ is the set of 3D model points, $m$ is the number of points in the set, $\boldsymbol{h}$ denotes the ground truth hand vertex vector, and $\tilde{\boldsymbol{h}}$ indicates the predicted hand vertex vector.

\subsection{Datasets}
We used the synthetic and real datasets presented by Hein, et al \cite{hein2021towards} for benchmarking the performance of HMD-EgoPose. Both datasets used the Colibri II battery powered orthopedic drill (DePuy Synthes, Raynham, MA, U.S) for rigid tool tracking and incorporated hand information while grasping the instrument. Images were annotated with the 6DoF tool pose of the instrument in frame and the 3D hand joints of the user grasping the tool.

\subsubsection{Synthetic Colibri dataset}
The synthetic Colibri (SynColibri) dataset included a total of $10,500$ image frames ($256 \times 256$ pixel resolution), created through a detailed synthetic data generation pipeline \cite{hein2021towards}. More details on the data generation pipeline are available from the authors \cite{hein2021towards}. 

Our framework was implemented using the PyTorch (\url{https://pytorch.org/}, accessed on 23 February 2022) deep learning library. We trained our network on the SynColibri dataset for $500$ epochs with a batch size of $32$. For all experiments, we used a computing workstation with an AMD Ryzen 9 3900X CPU and a single NVIDIA RTX 3090 GPU with 24 GB GDDR6X memory. For optimization on the SynColibri dataset, we used the Adam \cite{kingma2014adam} optimizer and an initial learning rate of $1e^{-4}$. During training on both the SynColibri and real Colibri (RealColibri) datasets, we performed extensive 6D augmentation of input data including random rotations, scaling, and translations as well as color space augmentation like adjusting contrast and brightness \cite{bukschat2020efficientpose}. For both the SynColibri and RealColibri dataset, we replicated the same five-fold cross-validation strategy as in \cite{hein2021towards} to assess variance across data splits and ensure our results are directly comparable to benchmarks. 

\subsubsection{Real Colibri dataset}
The RealColibri dataset included a total of $3,746$ frames ($256 \times 256$ pixel resolution) extracted from a total of $11$ individual recordings and captured through use of a mock operating room and human cadaveric specimen with an open incision \cite{hein2021towards}. 

We fine-tuned our network on the RealColibri dataset, using the best performing weights from network training on the SynColibri dataset, for $500$ epochs with a batch size of $32$. For optimization on the RealColibri dataset, we used stochastic gradient descent (SGD) with a momentum of $0.9$ and a reduced learning rate of $1e^{-5}$. We found the Adam optimizer led to faster convergence and better results on the SynColibri dataset; on the RealColibri dataset, the SGD optimizer with a reduced learning rate provided more stable convergence and results.

\subsection{Evaluation metrics}
We measured the mean 3D errors of our predicted rigid surgical drill pose relative to the ground truth pose ($\mathrm{ADD_{tool}}$) and provided assessments of positional error (in mm) and rotational error (in degrees) at the tip of the 3D drill model. We defined our measure of mean 3D errors of surgical drill pose as:
\begin{equation}
    \mathrm{ADD_{tool}} = \frac{1}{m}\sum \limits_{x\in M} \| (\mathrm{R}x + \boldsymbol{t}) - (\mathrm{\tilde{R}}x+\boldsymbol{\tilde{t}})\|\label{eq7}
    \end{equation}
where $m$ is the number of points in the 3D model set, $\mathrm{R}$ and $\mathrm{\tilde{R}}$, and $\boldsymbol{t}$ and $\boldsymbol{\tilde{t}}$ are the ground truth and predicted rotation matrices and translation vectors respectively. The 3D model points of the rigid surgical drill model set are represented by $x$. Additionally, we assessed the performance of our 3D hand vertex predictions by computing the mean end-point error across the $21$ predicted joints ($\mathrm{ADD_{hand}}$) \cite{hasson2020leveraging,hein2021towards}. We described our measure of hand vertex error as: 
\begin{equation}
    \mathrm{ADD_{hand}} = \frac{1}{m}\sum \limits_{x\in M} \| \boldsymbol{h}x - \boldsymbol{\tilde{h}}\|\label{eq8}
    \end{equation}
where $m$ is the number of points in the 3D set, and $\boldsymbol{h}$ and $\tilde{\boldsymbol{h}}$ are the ground truth and predicted hand vertex vectors respectively. To evaluate the runtime performance of our network, we included an estimate of the total number of parameters in each model, the model size, the FLOPS for an input image of size $(1,3,256,256)$, and the inference time (latency) measured using a single NVIDIA RTX 3090 GPU and a batch size of 1 across 1000 samples of the testing data. 

\subsection{Low-latency streaming for head-mounted display led pose estimation}
We used the Microsoft HoloLens 2 (\url{https://www.microsoft.com/en-us/hololens}, accessed on 23 February 2022) OST-HMD to record egocentric video frames and display 6DoF pose estimations of the tracked surgical drill as an augmented 3D virtual model. To estimate the required intrinsic parameters for computing 6DoF object pose, we performed a standard camera calibration procedure of the RGB photo-video camera sensor ($896 \times 504$ pixels). Intrinsic parameters including focal length ($f_x,f_y$), principal points ($p_x,p_y$), and skew ($s$) were computed to describe perspective projection to relate 3D points in the camera coordinate frame to their 2D image projections \cite{zhang2000flexible}. In our setup, the computing workstation relied on knowledge of the HoloLens 2 camera intrinsic parameters during computation of 6DoF object pose.

\subsubsection{Video streaming from head-mounted display to desktop}
We requested egocentric input video frames of size $896 \times 504$ pixels from the HoloLens 2 photo-video RGB camera at 30 frames per second (FPS). 
We built our application using the Unity C\texttt{\#} development platform (\url{https://unity.com/}, accessed on 23 February 2022) and implemented our video and data communication protocol using the WebRTC library (\url{https://github.com/microsoft/MixedReality-WebRTC}, accessed on 23 February 2022) to enable low-latency communication with our remote desktop peer. With a WebRTC C\texttt{++} dynamic link library (DLL), we used the universal datagram protocol (UDP) and an H.264 codec for streaming video frames in a compressed YUV420 representation from the HoloLens 2 to our high-performance computing workstation. 

\subsubsection{Optimized inference on a computing workstation}
The high-performance computing workstation included a WebRTC C\texttt{\#} Net Core project to receive and unpack the incoming YUV420 video frames from the HoloLens 2 to an RGB representation at the original $896 \times 504$ pixels and 30 FPS representation. We used a C\texttt{++} DLL and OpenCV (\url{https://opencv.org/}, accessed on 23 February 2022) within the C\texttt{\#} project for optimal pre-processing of the incoming video frames, including image normalization and bilinear interpolation to a size of $256 \times 256$ pixels. For optimal GPU inference, we used the NVIDIA TensorRT software development kit (\url{https://developer.nvidia.com/tensorrt}, accessed on 23 February 2022) and ONNX library (\url{https://onnx.ai/}, accessed on 23 February 2022). Following network inference, filtering of the resulting predictions included rescaling, clipping and non-maximum suppression to arrive at a final 6DoF pose prediction result. Using a WebRTC data channel, we sent the resulting axis-angle rotation and translation vectors as a combined 24-byte array to the HoloLens 2 device.

\section{Results and Discussion}\label{sec3}
In this section, we evaluated the performance of HMD-EgoPose relative to current state-of-the-art approaches and assessed the performance and feasibility of 6DoF pose estimation on a commercially available OST-HMD through our low-latency streaming approach and computing workstation. 

\subsection{Experimental results} 
Hein et al. have presented several strategies for surgical drill and hand pose estimation from monocular RGB data for the synthetic drill dataset based off of the PVNet \cite{peng2019pvnet} and HandObjectNet \cite{hasson2020leveraging} frameworks. PVNet focuses solely on 6DoF object pose estimation and does not consider the joint interaction of the hand of a user with the object \cite{peng2019pvnet}. PVNet indirectly estimates object pose using a set of 2D keypoints that correspond to the 3D bounding box center and selected 3D locations across the surface of the model. To estimate a segmentation mask and 2D vector field for each keypoint, PVNet uses a similar architecture as a U-Net \cite{ronneberger2015u}. A RANSAC-based voting scheme is used to recover 2D keypoints from their respective 2D vector fields. The final 6DoF tool pose is recovered using a perspective-n-point (PnP) approach to minimize the Mahalanobis distance based on the mean and covariance of the keypoint predictions and the ground truth keypoints. Instead of solely focusing on 6DoF object pose alone, HandObjectNet considers the joint interaction of the hands of a user and the surgical drill \cite{hein2021towards}. HandObjectNet uses a shared ResNet-18 \cite{he2016deep} encoder and decoders for both the hand and object \cite{hasson2020leveraging}. The hand decoder branch estimates 18 pose parameters, consisting of 15 coefficients to describe the hand configuration and 3 parameters to detail the axis-angle representation of the hand, and 10 shape parameters to describe the MANO hand model \cite{romero_embodied_2017}. The object branch regresses an axis-angle rotation vector, 2D translation vector, and a focal-normalized depth offset; the 3D translation is then composed using knowledge of the camera intrinsic parameters \cite{hasson2020leveraging}. 

\subsubsection{Synthetic Colibri dataset}
\begin{figure}[h]%
    \centering
    \includegraphics[width=0.95\textwidth]{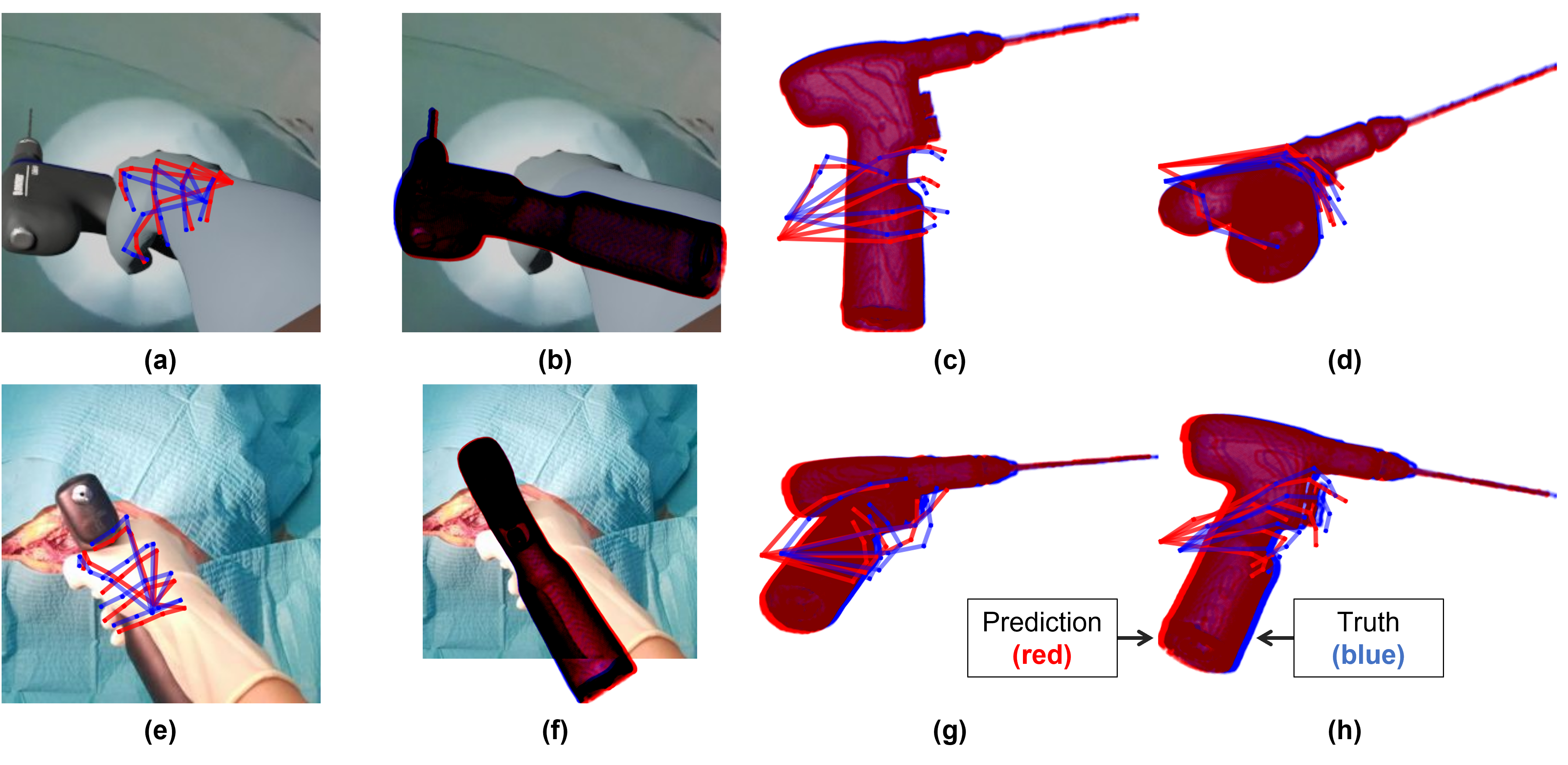}
    \caption{Sample three-dimensional (3D) hand vertex and six-degree-of-freedom (6DoF) object pose estimates are shown relative to ground truth labels for image samples from the synthetic Colibri (a-d) and real Colibri (e-h) datasets. Predicted labels for hand vertices and model pose are shown in red, and ground truth labels for hand vertices and model pose are in blue. The 2D projections of 3D hand vertices (a, e) and 2D projections of 3D model points (b, f) are shown on the related input image frame. A view from the top (c, g) and the right (d, h) are included to show the 3D model and hand vertex prediction agreement from different view perspectives.}\label{fig2}
    \end{figure}

Table \ref{tab1} compares the performance and latency of our approach relative to other state-of-the-art techniques on the SynColibri surgical drill dataset. Our HMD-EgoPose framework outperformed the HandObjectNet \cite{hein2021towards,hasson2020leveraging} and PVNet \cite{hein2021towards, peng2019pvnet} techniques in measures of tool ADD and rotational error at the drill bit tip. Our approach used $3 \times$ fewer network parameters, required $3 \times$ less memory for deployment, achieved $3 \times$ faster FLOPS, and required less time for inference than HandObjectNet, the previous best performing approach. Across the synthetic dataset, we achieved an average tool 3D vertex ($\mathrm{ADD_{tool}}$) accuracy of 11.17 mm. Figure \ref{fig2}(a-d) includes a sample image frame from the SynColibri dataset with corresponding network predictions for 3D hand vertices and 6DoF rigid surgical drill pose as compared to the ground truth labels.

\begin{table}[h]
\begin{center}
\begin{minipage}{\textwidth}
\caption{Networks trained and evaluated on the synthetic Colibri surgical drill dataset (mean $\pm$ standard deviation). We have directly included the tool ADD, hand ADD, drill tip error, and drill tip direction error results reported by Hein et al. for comparison to our approach \cite{hein2021towards}.}\label{tab1}%
\begin{tabular}{@{}lccc@{}}
\toprule
 & HMD-EgoPose & HandObjectNet \cite{hein2021towards} & PVNet \cite{hein2021towards}\\
\midrule
Tool ADD (mm)    & $\boldsymbol{11.17 \pm 9.17} $   & $16.73\pm 16.97$  & $20.59 \pm 52.14$  \\
Drill Tip Error (mm)    & $34.81 \pm 36.01$   & $44.45 \pm 59.72$  & $\boldsymbol{31.10 \pm 67.18}$  \\
Drill Bit Direction Error (deg)    &  $\boldsymbol{5.40 \pm 6.98}$  & $6.59 \pm 10.18$  & $7.11 \pm 21.78 $ \\
Hand ADD (mm)    & $19.79 \pm 7.72$  & $\boldsymbol{17.15 \pm 10.58}$  & \textemdash  \\
Network Paramters (M)    & $\boldsymbol{3.92}$   & $12.49$  & $12.96$  \\
Network Size (MB)    & $\boldsymbol{16.3}$   & $53.1$  & $51.9$  \\
FLOPS (B)    & $\boldsymbol{1.38}$   & $4.76$  & $30.96$  \\
Latency (ms)    & $\boldsymbol{19.8 \pm 2.3}$   & $21.5 \pm 3.3$  & $52.4 \pm 8.2$  \\
\botrule
\end{tabular}
\end{minipage}
\end{center}
\end{table}

\subsubsection{Real Colibri dataset}
Table \ref{tab2} describes the performance and latency of our approach relative to other state-of-the-art techniques on the RealColibri surgical drill dataset. Our HMD-EgoPose framework outperformed the HandObjectNet \cite{hein2021towards,hasson2020leveraging} and PVNet \cite{hein2021towards,peng2019pvnet} techniques in measures of tool ADD and rotational and translational error at the drill bit tip. Across the real dataset, we achieved an average tool 3D vertex ($\mathrm{ADD_{tool}}$) accuracy of 11.00 mm. Figure \ref{fig2}(e-h) shows a sample image frame from the real Colibri dataset with corresponding network predictions for 3D hand vertices and 6DoF rigid surgical drill pose as compared to the ground truth labels. Even with significant mutual occlusion of the hand of a user and the surgical drill, there was good agreement between the 6DoF tool pose and the ground truth information due to the jointly learned hand and tool representation of HMD-EgoPose. Table \ref{tab3} indicates the performance of our proposed HMD-EgoPose approach when (1) trained and evaluated solely on the RealColibri dataset (no pre-training on the SynColibri dataset); and (2) trained solely on the SynColibri dataset and evaluated on the RealColibri dataset.

\begin{table}[h]
    \begin{center}
    \begin{minipage}{\textwidth}
    \caption{Networks pretrained on synthetic data and finetuned and evaluated on real Colibri surgical drill dataset (mean $\pm$ standard deviation). We have directly included the tool ADD, hand ADD, drill tip error, and drill tip direction error results reported by Hein et al. for comparison to our approach \cite{hein2021towards}.}\label{tab2}%
    \begin{tabular}{@{}lccc@{}}
    \toprule
     & HMD-EgoPose & HandObjectNet \cite{hein2021towards} & PVNet \cite{hein2021towards}\\
    \midrule
    Tool ADD (mm)    & $\boldsymbol{11.00 \pm 7.27} $   & $13.78\pm 5.28$  & $39.72 \pm 66.49$  \\
    Drill Tip Error (mm)    & $\boldsymbol{28.07 \pm 19.81}$   & $66.11 \pm 26.91$  & $72.80 \pm 105.66$  \\
    Drill Bit Direction Error (deg)    &  $\boldsymbol{3.89 \pm 2.74}$  & $8.71 \pm 3.98$  & $13.41 \pm 33.78 $ \\
    Hand ADD (mm)    & $17.35 \pm 8.09$  & $\boldsymbol{9.78 \pm 4.54}$  & \textemdash  \\
    \botrule
    \end{tabular}
    \end{minipage}
    \end{center}
    \end{table}

\begin{table}[h]
    \begin{center}
    \begin{minipage}{\textwidth}
    \caption{HMD-EgoPose performance when trained on the RealColibri or SynColibri dataset only and evaluated on the RealColibri dataset (mean $\pm$ standard deviation).}\label{tab3}%
    \begin{tabularx}{\textwidth}{>{\hsize=0.375\hsize}X
        >{\hsize=0.3125\hsize}C
        >{\hsize=0.3125\hsize}C}    
    \toprule
    HMD-EgoPose performance & Trained/Evaluated on RealColibri Only & Trained on SynColibri Only, Evaluated on RealColibri\\
    \midrule
    Tool ADD (mm)    & $17.16 \pm 11.01$  & $53.77 \pm 49.86$  \\
    Drill Tip Error (mm)   & $41.76 \pm 27.59$  & $251.38 \pm 237.93$  \\
    Drill Bit Direction Error (deg)  & $5.58 \pm 3.68$  & $37.54 \pm 41.20 $ \\
    Hand ADD (mm)    & $19.47 \pm 10.20$  & $63.06 \pm 33.62$  \\
    \botrule
    \end{tabularx}
    \end{minipage}
    \end{center}
    \end{table}
    
\subsection{Feasibility assessment of streaming to a head-mounted display}
\begin{figure}[h]%
    \centering
    \includegraphics[width=0.95\textwidth]{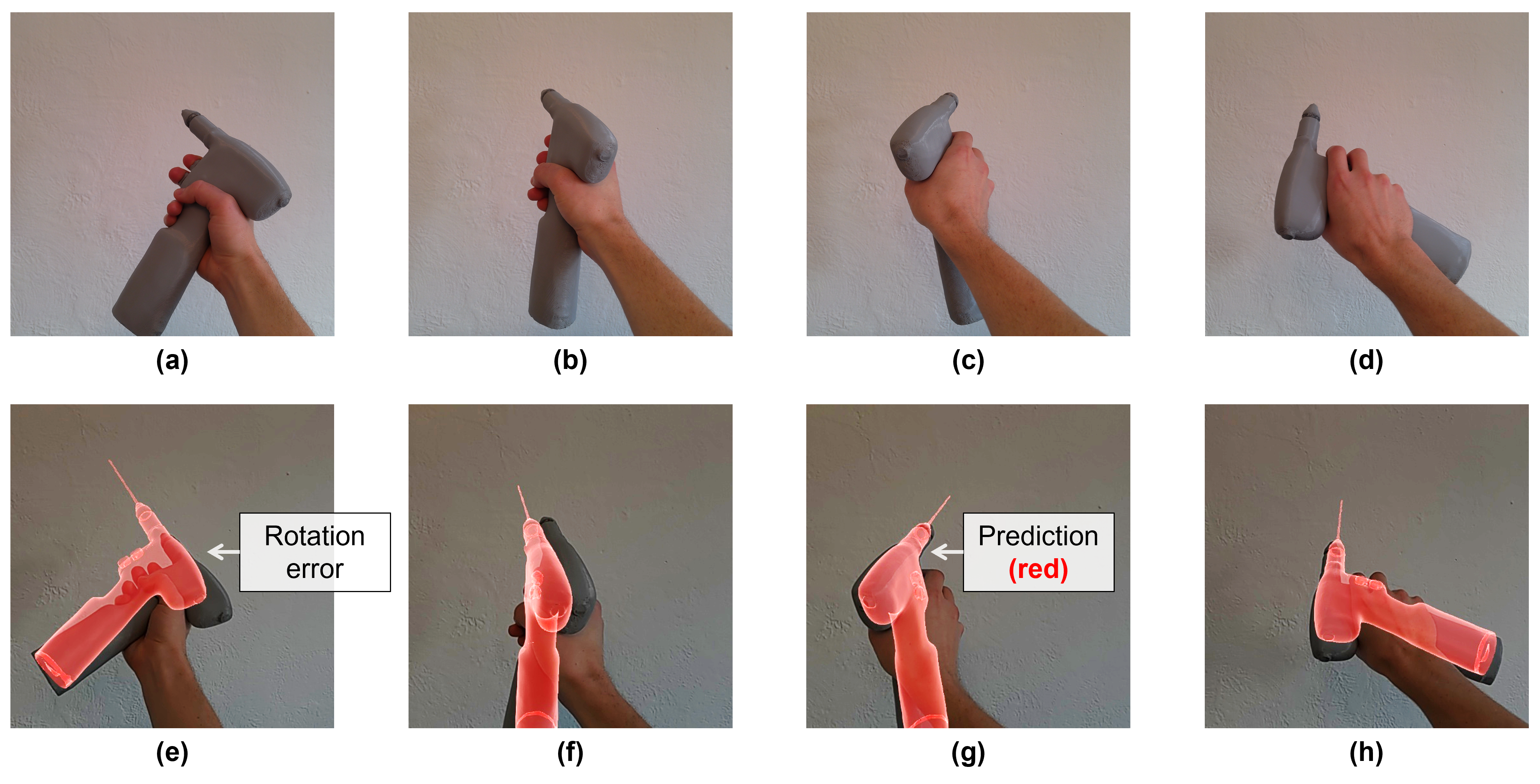}
    \caption{Sample images demonstrating the real-world tracking accuracy of our network for rigid surgical drill pose estimation on the HoloLens 2 head-mounted display while using a representative three-dimensional (3D) printed model with identical geometry as the Colibri II surgical drill. (a-d) Images captured using the HoloLens 2 photo-video camera prior to launching our tracking application. We include four separate grasping poses of the 3D printed surgical drill model. (e-h) Single frames replicating the grasping poses from images (a-d) captured using the HoloLens 2 mixed reality capture capabilities after launching our tracking application. The two-dimensional (2D) views are created by the mixed reality capture application using the right eye camera of the 3D stereo view on the headset, which is known to introduce an offset in the position of virtual models. Pose predictions from our HMD-EgoPose network are applied to the 3D surgical drill model displayed in red.}\label{fig3}
    \end{figure}

In Figure \ref{fig3}, we include a set of sample images from an experiment where a user was asked to hold the 3D printed surgical drill model and move the drill as if they were replicating the movements typical to its use in an orthopedic procedure. As indicated in the sample frames, there was good agreement between both the predicted and ground truth surgical drill pose estimates. We assessed the mean 2D registration accuracy between the 3D printed surgical drill model and the 3D virtual model visually by capturing $25$ still-frame images and measuring the error at three landmark positions: the drill chuck (where the drill bit attachment would be inserted), the base of the drill handle, and the circle on the top back surface of the drill just above where a user would grasp. We assessed the mean 2D Euclidean error at these three landmark positions across the image set as $37.97 \pm 26.41$ pixels (mean $\pm$ standard deviation).

In our feasibility assessment, the network was used exclusively by the HoloLens 2 and the computing workstation. Across 25 measurements, we computed the average available network bandwidth as $3.17 \pm 1.75$ Mbps and network latency as $4.84 \pm 1.83$ ms between the HoloLens 2 and the computing workstation over our local Wi-Fi connection, with a peak measured bitrate of $8.04$ Mbps.

Across 100 consecutive image frames, we characterized the average pixel-to-photon latency -- the cumulative time required for streaming of egocentric video to our high-performance computing workstation, performing GPU inference, returning the results to the HoloLens 2, and rendering them on-screen -- as $199.1 \pm 30.3$ ms (roughly 5 FPS). Of this, roughly 160 ms was required for video frame transmission, 6 ms for frame resizing and rescaling, 12 ms for GPU inference in TensorRT and post prediction filtering, 8 ms for return transmission of the predicted 6DoF transformation to the HoloLens 2, and 16 ms for rendering of the updated virtual model pose. Additional qualitative videos demonstrating the performance of HMD-EgoPose during network inference are available in the supplementary material. Importantly, we used a threading approach to offload the update of 6DoF pose results off of the main rendering thread so as not to slow the HoloLens 2 app operation to 5 FPS. With our approach, we achieved rendering performance at 50-60 FPS, video capture at 30 FPS, and update of the 6DoF pose predictions at 5 FPS.

To assess the latency of our network when run directly on the computing hardware of the HoloLens 2: the on-board CPU, we performed an identical experiment and measured a pixel-to-photon latency of $1097.54 \pm 220.21$ ms across 100 consecutive frames. Unfortunately, due to limited GPU support on the HoloLens 2, it is more efficient to perform CPU-based inference \cite{doughty2021surgeonassist}. Though our HMD-EgoPose model is compact in terms of size and FLOPS, many of the network layers and operations of the BiFPN are inefficient to perform on a CPU, contributing to the slow on-device performance of roughly 1 FPS.

\subsection{Limitations}
As we were unfortunately not able to access the specific Colibri II surgical drill model used in collection of the SynColibri benchmark dataset, we created a representative 3D printed surgical drill model of identical geometry to use for qualitative evaluation. Even though our 3D printed model did not have a similar color or texture scheme as the real drill, we were able to demonstrate promising tracking performance using our HMD-EgoPose network. As is visible in the supplementary video material, due to the heavy cropping of input training data to center on the surgical drill, our network loses tracking when the drill moves away from the center of the image frame. A future fix for this problem could be to train a lightweight object detection framework for prediction of a 2D bounding box around the surgical drill for use in adaptive cropping of input video frames.

Another limitation of this work is with regard to the tracking latency and performance. In our current implementation, we achieved a pixel-to-photon latency of 199.1 ms, resulting in tracking performance of roughly 5 FPS. We set out to optimize the latency of image frame preprocessing and GPU-based inference; however, the main contributor to latency was the roughly 160 ms that was required for image data transmission from the HoloLens 2 to the computing workstation. We anticipate that there could have been further reductions to network latency by reducing frame size to the desired $256 \times 256$ pixels on the HoloLens 2 prior to transmission to reduce packet size. Over 100 consecutive measurements, we quantified the mean time for frame resizing on the HoloLens 2 using nearest neighbour interpolation as $44.04 \pm 28.49$ ms, meaning that any reduction in transmission time would need to be greater than $44.04$ ms to improve the overall pixel-to-photon latency.


\section{Conclusion}
The focus of this work was to contribute to the improvement of egocentric 6DoF object and hand pose estimation strategies from monocular RGB data through further investigating the mutual interactions of the hand of a user and a surgical instrument. With HMD-EgoPose, we demonstrated state-of-the-art performance in hand and surgical drill pose estimation as well as network inference speed through our single-shot pose estimation approach. Further, we showed the feasibility of a low-latency streaming method to enable AR-based egocentric surgical navigation on a commercially available OST-HMD, opening numerous possible avenues for use in surgical tool tracking, 3D to 3D OST-HMD display calibration, and tracking of other rigid landmark objects in the operating room.

Our HMD-EgoPose framework for egocentric surgical tool and hand tracking provides researchers with the means for improved tracking accuracy and robustness to occlusion during challenging 3D tasks in AR-led surgical navigation and, with future developments to the network architecture and training dataset diversity, could contribute to improved patient outcomes. With modifications to the training dataset for the specific intervention, we expect that our modular framework could have significant implications in many other surgical interventions which rely on the tracking of surgical tools or instruments for guidance.












\noindent


\bibliography{sn-bibliography}


\end{document}